\documentclass[journal,twoside,web]{ieeecolor}
\usepackage{generic}
\usepackage{cite}
\usepackage{amsmath,amssymb,amsfonts}
\usepackage{algorithm,algorithmic}
\usepackage{graphicx}
\usepackage{textcomp}
\usepackage{tabularx,array,booktabs,arydshln}
\usepackage{url}
\usepackage{etoolbox}
\providecommand{\refname}{References}
\patchcmd{\thebibliography}
    {\itemsep 0pt plus pt\relax}
    {\itemsep 0pt\relax}
    {}
    {\PackageError{stsg-arxiv}{Bibliography spacing patch failed}{Check ieeecolor.cls.}}
\usepackage{hyperref}
\hypersetup{
    hidelinks,
    pdftitle={STSG-VQA: Evidence-Grounded Temporal Question Answering from Surgical Spatio-Temporal Scene Graphs},
    pdfauthor={Jing Li and Duygu Sarikaya},
    pdfsubject={Surgical Video Question Answering},
    pdfkeywords={Surgical Video Question Answering, Spatio-temporal Scene Graphs, Temporal Reasoning}
}
\def\BibTeX{{\rm B\kern-.05em{\sc i\kern-.025em b}\kern-.08em
    T\kern-.1667em\lower.7ex\hbox{E}\kern-.125emX}}

\graphicspath{{imgs/}}

\makeatletter
\let\STSG@originaltitlepagestyle\ps@titlepagestyle
\def\ps@titlepagestyle{%
    \STSG@originaltitlepagestyle
    \def\@oddhead{\hbox{}\scriptsize\sffamily\leftmark\hfil\thepage}%
    \def\@evenhead{\scriptsize\sffamily\thepage\hfil\leftmark\hbox{}}%
}
\makeatother

\begin{document}
\title{STSG-VQA: Evidence-Grounded Temporal Question Answering from Surgical Spatio-Temporal Scene Graphs}

\author{%
    \makebox[\linewidth][c]{Jing~Li and Duygu~Sarikaya}%
    \thanks{Jing Li and Duygu Sarikaya are with the
    School of Computer Science, University of Leeds,
    UK
    (E-mail: \{sc232jl, D.Sarikaya\}@leeds.ac.uk).}%
}

\maketitle

\begin{abstract}
\label{sec:abstract}
    Despite recent advances in surgical vision-language models (VLMs), temporal reasoning remains limited because existing supervision is largely frame-centric. Frame-level scene graphs (SGs) have proven effective in providing structured representations of surgical environments but do not explicitly model the dynamics of surgical workflows. To explicitly model how surgical states evolve across time, we introduce a multi-level structured temporal supervision methodology that augments frame-level surgical SGs with object-level continuity, event-level interaction continuity, and procedure-level connectivity. We then execute temporal queries over the resulting spatio-temporal scene graphs (STSGs) to generate evidence-grounded question--answer pairs, which together form the STSG-VQA benchmark. Each question is linked to the temporal interval and STSG evidence used to derive its reference answer, enabling traceable verification. The benchmark contains 18,458 question--answer pairs across seven temporal categories. Fine-tuning Qwen3-VL-4B and Hulu-Med-4B with STSG-derived supervision improves question-level micro accuracy by 24.39 and 19.56 percentage points over their zero-shot baselines and by 16.50 and 14.25 points over static scene-graph supervision, respectively. These gains span all temporal categories, indicating that STSG-derived supervision helps surgical VLMs reason over temporally grounded interactions rather than isolated frames. The code and dataset will be made publicly available upon acceptance.
    
\end{abstract}


\section{Introduction}
\label{sec:introduction}

    \IEEEPARstart{D}{riven} by recent breakthroughs in vision-language models (VLMs), surgical visual question answering (VQA) has achieved remarkable progress, demonstrating great promise for decision support, postoperative analysis, and surgical education. Surgical videos encompass rich and highly structured procedural information, characterized by transitions between surgical phases, the entry and exit of surgical instruments within the field of view, and the deformation of anatomies under traction. Fundamentally, clinically meaningful surgical actions are typically defined by the interactions among instruments, actions, and targets, formulated as an action triplet $\langle\text{instrument}, \text{verb}, \text{target}\rangle$ \cite{tripnet}. Therefore, a practical surgical VLM should not be confined to single-frame content recognition. Rather, it requires robust temporal understanding to answer dynamic questions, including tracking object movements, capturing ongoing interactions, and detecting contextual changes before and after phase transitions. However, developing VLMs tailored for surgical applications poses unique challenges. This is primarily attributed to the scarcity of domain-specific datasets, particularly those assessing temporal understanding and reasoning capabilities.

    \begin{figure}[!t]
    \centering
    \includegraphics[width=\columnwidth]{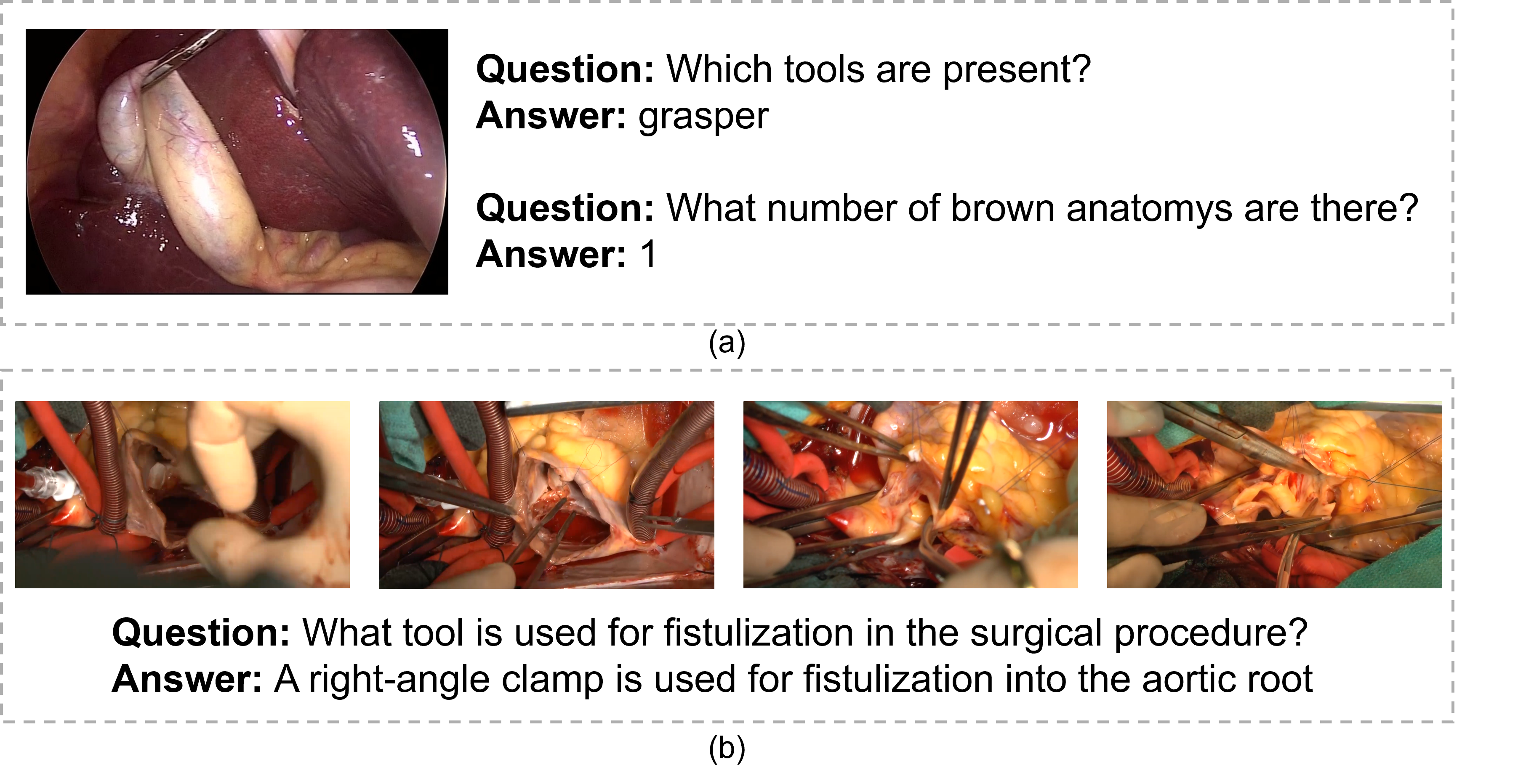}
    \caption{Two forms of limited temporal supervision in existing surgical VQA datasets. (a) Static questions focus on visible instrument identity and anatomical counting \cite{ssgvqa}. (b) A multi-frame example asks for a procedure-associated instrument and does not explicitly require reasoning over temporal relations among the frames \cite{surgpubvideo}. The question wording and answers are reproduced verbatim from the original datasets.}
    \label{fig:ssgvqa_surgpubvideo}
    \end{figure}

    Improving temporal reasoning requires more than exposing a model to additional video frames; the supervision must explicitly capture how surgical states interact and change over time. This motivates a structured temporal representation that enables models to learn how surgical scenes evolve rather than only what is visible in individual frames. However, existing surgical VQA datasets provide limited support for fine-grained temporal supervision. As illustrated in Fig.~\ref{fig:ssgvqa_surgpubvideo}, some datasets provide only frame-level supervision (a), whereas others include video clips but pose coarse-grained recognition questions that do not explicitly probe temporal relations (b). Moreover, question wording and procedural priors may enable models to answer without sufficiently relying on visual evidence \cite{surgcheck}.
  
    Recent studies have increasingly incorporated temporal encoders and video-language training objectives \cite{surgvivqa,surgtemp}, and large surgical VLMs have expanded the range of surgical visual tasks that can be addressed with instruction tuning \cite{surgvlm}. Nevertheless, the supervision and evaluation signals for fine-grained temporal reasoning remain limited. In particular, correctly answering a coarse-grained recognition question from one or a few salient frames does not establish that a model can track the same surgical entities across time or reason about how their interactions persist and change throughout a procedure.
    
    Structured representations offer a complementary path. Scene graphs encode objects and their relations, and have been used to improve compositional visual reasoning in both general \cite{actiongenome} and surgical \cite{ssgvqa, orgraph} settings. In surgery, scene-graph-based VQA is particularly well suited because instrument-target interactions are naturally expressed as structured triplets. However, frame-level surgical scene graphs represent objects and relations within individual images \cite{ssgvqa} and do not explicitly associate corresponding object instances or interaction triplets across frames. Without such cross-frame connectivity, repeated observations of the same triplet remain independent frame-level facts rather than being consolidated into a temporally bounded event. Consequently, questions generated directly from these graphs are limited to frame-local properties.
    
    To address these limitations, we introduce a multi-level structured temporal supervision methodology for surgical video reasoning. The methodology transforms isolated frame-level scene graphs (SGs) \cite{ssgvqa} into an event-centered representation that jointly models object persistence and interaction continuity. We instantiate this methodology through spatio-temporal scene graphs (STSGs) and an evidence-grounded temporal query framework, resulting in the STSG-VQA benchmark. It covers seven question categories: count, duration, ordering, extreme, boundary, phase transition, and concurrency. In total, it comprises 18,458 QA pairs split into training, validation, and test sets of 13,932, 1,775, and 2,751, respectively. The evaluation protocol supports deterministic answer formats and LLM-judged (Qwen3.5-4B \cite{qwen3.5}) open-ended interaction descriptions. Initial baselines with Qwen2.5-VL \cite{qwen25vl}, Qwen3-VL \cite{qwen3vl}, and LLaVA-NeXT-Video \cite{llavanextvideo} show that current VLMs remain weak on such surgical temporal understanding and reasoning tasks. Our contributions and findings are below:
    \begin{itemize}

        \item We introduce an event-centered structured temporal reasoning methodology that extends frame-level surgical scene graphs, transforming isolated surgical scenes into a representation of their temporal evolution.

        \item We introduce an evidence-grounded temporal query framework that operationalizes the proposed representation into seven reasoning capabilities: event counting, duration aggregation, interaction ordering, event concurrence, dominant-activity identification, immediate cross-boundary reasoning, and broader pre/post procedural-change reasoning.

        \item We proposed the STSG-VQA, a benchmark containing 18,458 question--answer pairs from 45 laparoscopic videos \cite{rendezvous,cholectsplits,ssgvqa}, together with a unified evaluation protocol covering deterministic and open-ended answers.

        \item We show that structured temporal supervision benefits both temporal reasoning and static surgical VQA, increasing the contribution of visual evidence, and yielding the strongest frame-level performance when combined with static supervision.

    \end{itemize}

\section{Related Work}
\label{sec:related_work}

\subsection{Surgical Vision-Language Models}
    The rise of VLMs has demonstrated remarkable capabilities in medical image understanding. CLIP-style models \cite{medclip, pubmedclip, biomedclip, surgvlp} learn joint visual-textual representations by aligning medical images with their corresponding clinical reports, thereby supporting downstream applications such as automated report generation and image-to-text retrieval. 

    Building upon these vision–language alignment frameworks, multimodal large language models such as LLaVA-Med \cite{llavamed}, Med-Flamingo \cite{medflamingo}, and Med-Gemini \cite{medgemini} have extended medical image understanding toward instruction following, VQA, and open-ended multimodal dialogue. In surgery, domain-specific models such as SurgVLM  \cite{surgvlm} and SurgVidLM \cite{surgvidlm} have further extended this paradigm toward surgical perception, temporal analysis, and video-level understanding. These models operate as multimodal reasoning systems that integrate imaging data, patient records, and clinical guidelines within unified medical workflows, enabling more context-aware clinical understanding.
 
    Although surgical VLMs have achieved promising results on image-based tasks, extending them to video understanding remains challenging because models must reason over complex temporal dependencies across evolving surgical scenes \cite{surgicalsceneunderstandingreview}. This requires models to identify transient visual evidence, integrate instrument-tissue interactions across frames, and retain procedural context over extended temporal intervals \cite{surgvivqa, surgtemp, surgvidlm}. 

\subsection{Surgical Visual Question Answering}
    Surgical VQA has emerged as a specialized research paradigm that extends conventional VQA frameworks to the surgical domain. By jointly modeling visual information and natural language queries in complex operative scenes, it offers a promising direction for clinical training and decision support \cite{surgtemp}. Early work introduced transformer-based surgical VQA on endoscopic images \cite{surgicalvqa}, while Surgical-VQLA extended the task by jointly predicting an answer and a bounding box around the question-relevant visual region in robotic surgery \cite{surgicalvqla}. However, this grounding remains spatial and frame-level, without identifying how the corresponding entities or interactions persist across time. PitVQA further explored LLM-based VQA for pituitary surgery with image-grounded text embeddings \cite{pitvqa}. More recently, SSG-VQA incorporated surgical scene graph knowledge to reduce question-conditioned bias and support geometry-aware reasoning over surgical triplets \cite{ssgvqa}. 
    
    SurgViVQA advanced video-level surgical VQA by introducing a Masked Video–Text Encoder that fuses questions with short video clips to learn temporally aware latent representations for open-ended surgical question answering \cite{surgvivqa}. While this advances video-level surgical VQA beyond isolated-frame analysis, temporal evidence remains encoded in latent clip-level representations rather than explicitly represented through object tracks or temporally bounded interaction events. Our work instead introduces an event-centered STSG representation and derives evidence-grounded supervision for temporal reasoning.

\subsection{Structured Knowledge for Enhanced Reasoning}
    Structured visual knowledge has been used to support reasoning beyond visual features. Visual Genome introduced dense object, attribute, and relationship annotations for image-level SG reasoning \cite{visualgenome}. Action Genome extended this idea to videos by representing actions as spatio-temporal scene graphs, i.e., temporally ordered object-relation graphs that capture changes occurring throughout an action \cite{actiongenome}. Several studies \cite{hostsg, vot, step} have shown that explicitly representing objects across consecutive frames and their relationships as nodes and edges can be effectively applied to video understanding tasks. In surgical video analysis, structured labels such as triplets have been shown to capture fine-grained tool-tissue interactions \cite{rendezvous}, and recent surgical scene graph work models objects, actions, and anatomical relations for downstream understanding \cite{ssgvqa, holistic}. However, such representations remain largely underexplored in surgical video analysis, where frequent instrument-tissue occlusions, rapid camera motion, and abrupt appearance changes (e.g., smoke, blood, trocar views) make cross‑frame identity association unstable \cite{surgvlm}.

    \begin{figure*}[!t]
    \centering
    \includegraphics[width=\textwidth]{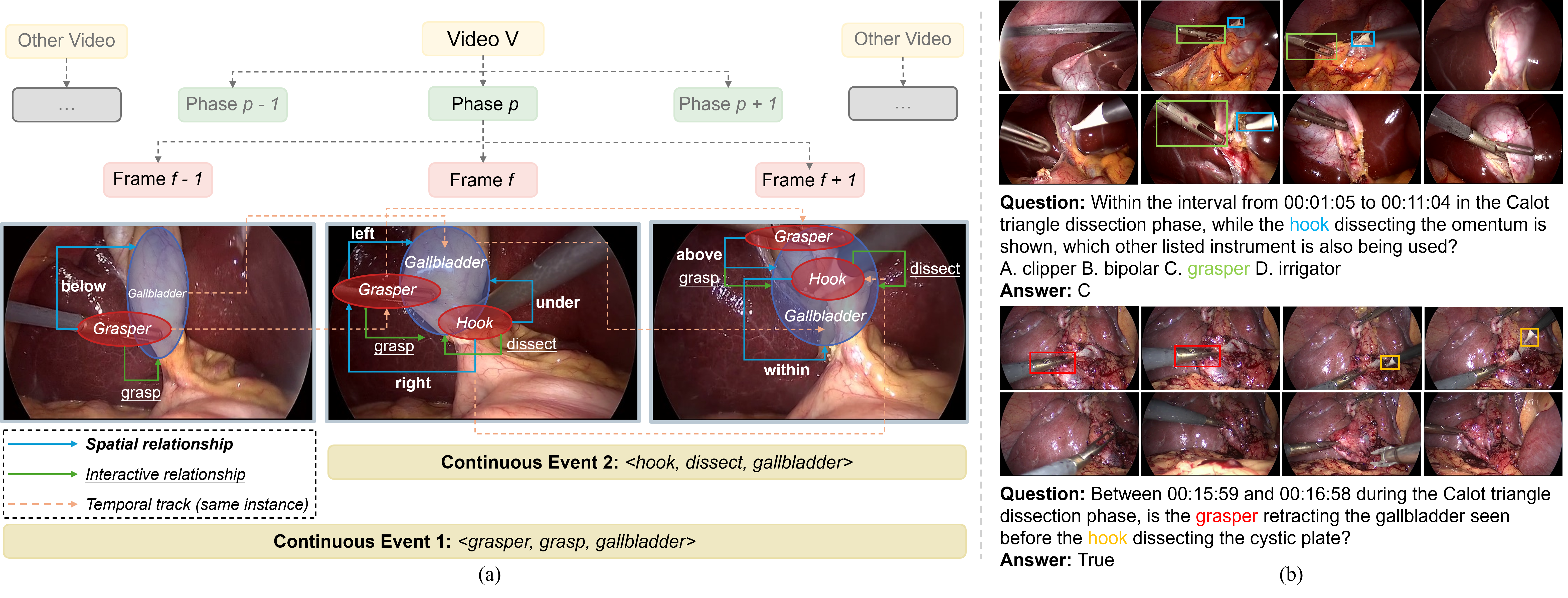}
    \caption{(a) Overview of STSG construction from frame-level surgical scene graphs. Within-frame spatial and instrument–target interaction edges preserve local semantic relations, temporal tracks link observations of the same object across frames, and repeated triplets are aggregated into continuous event nodes. Line styles denote spatial, interaction, and temporal associations, as indicated in the legend. (b) Representative temporal QA instances. The top sample illustrates a concurrency question, which assesses the co-occurrence of surgical interactions by identifying another instrument being used while the hook dissects the omentum. The bottom example illustrates an ordering question, which assesses the temporal precedence between the grasper retracting the gallbladder and the hook dissecting the cystic plate. Bounding boxes and highlighted object instances indicate the visual evidence associated with the corresponding objects and interactions.}
    \label{fig:STSG_with_2_QAs}
    \end{figure*}

\section{Methodology}
\label{sec:methodology}

    \begin{table}[!t]
    \caption{Comparison with representative surgical VQA benchmarks.}
    \label{tab:dataset_comparison}
    \centering
    \footnotesize
    \setlength{\tabcolsep}{2.2pt}
    \renewcommand{\arraystretch}{1.12}
    \resizebox{\columnwidth}{!}{
    \begin{tabular}{lcccc}
    \hline
    \textbf{Dataset} &
    \textbf{Visual scope} &
    \textbf{Answer format} &
    \textbf{Temporal reasoning} &
    \textbf{Structured evidence} \\
    \hline
    EndoVis18-VQA &
    Frame image &
    Closed-set &
    No &
    No \\
    
    PitVQA &
    Frame image &
    Open-ended &
    No &
    No \\
    
    SSG-VQA &
    Frame-level SG &
    Closed-set &
    No &
    SG \\
    
    REAL-Colon-VQA &
    8-frame clip &
    Open-ended &
    Yes &
    No \\
    
    STSG-VQA (ours) &
    Video segment &
    Both &
    Yes &
    STSG \\
    \hline
    \end{tabular}
    }
    \end{table}

\subsection{Overview}
    We formulate surgical temporal reasoning through three complementary forms of structured connectivity: object-level continuity, which preserves entity identity across frames; event-level continuity, which consolidates repeated interactions into temporally extended surgical events; and procedure-level connectivity, which relates these events to surgical phases. Together, these components transform isolated frame-level scene descriptions into an event-centered representation of surgical state evolution. We use this representation to derive evidence-grounded temporal supervision, with each question and answer traceable to the graph relations and temporal intervals from which it is generated, rather than relying solely on free-form natural-language annotation. As summarized in Table~\ref{tab:dataset_comparison}, STSG-VQA differs from existing surgical VQA benchmarks by jointly supporting video-level temporal reasoning and explicit evidence grounding through the STSG representation.

\subsection{Multi-Level Spatio-Temporal Representation}

    We model each surgical video as an ordered sequence of frame-level SGs and construct the corresponding video-level STSG by introducing temporal, procedural, and event-level connectivity, as illustrated in Fig.~\ref{fig:STSG_with_2_QAs} (a). For video $i$, the SG at frame $t$ is represented as
        \begin{equation}
            \mathrm{SG}_{i,t}
            =
            \left(
            \mathcal{O}_{i,t},
            \mathcal{R}_{i,t}
            \right),
            \label{eq:frame_sg}
        \end{equation}
    where $\mathcal{O}_{i,t}$ denotes the set of detected entities and $\mathcal{R}_{i,t}$ denotes the set of intra-frame semantic relations. Each object in $\mathcal{O}_{i,t}$ stores its basic visual and semantic attributes (e.g., name, entity type, bounding box). The relation set contains spatial predicates between detected entities and surgical action triplets, represented as $\langle \mathrm{instrument}, \mathrm{verb}, \mathrm{target} \rangle$, which encode the acting instrument, the performed surgical action, and the corresponding target, respectively.

    The video-level STSG is obtained by composing the phase-aware frame-level SGs and subsequently augmenting them with cross-frame temporal links and event-level abstractions:
        \begin{equation}
            \mathrm{STSG}_i
            =
            \operatorname{Compose}
            \left(
            \left\{
            \left(
            \mathrm{SG}_{i,t},p_i(t)
            \right)
            \right\}_{t\in\mathcal{T}_i}
            \right)
            \oplus
            \Delta_i^{\mathrm{temp}}
            \oplus
            \Delta_i^{\mathrm{evt}},
            \label{eq:video_stsg}
        \end{equation}
    where $\mathcal{T}_i$ denotes the chronologically ordered frame set of video $i$, $p_i(t)$ is the surgical phase associated with frame $t$, and $\oplus$ denotes graph augmentation with additional nodes or edges. The composition operator preserves the entity nodes and intra-frame relations of each SG while associating the corresponding frame with its procedural phase. The temporal linking $\Delta_i^{\mathrm{temp}}$ establishes object-level continuity across frames, whereas the event aggregation $\Delta_i^{\mathrm{evt}}$ abstracts persistent frame-level triplets into temporally grounded event nodes. The resulting STSG therefore provides a compact structured representation of the surgical video that emphasizes task-relevant entities, interactions, procedural context, and their temporal evolution.

    The temporal augmentation of the frame-level SGs proceeds in three stages: cross-frame object association, frame-level action grounding, and event-level temporal aggregation. These stages progressively transform isolated frame observations into persistent entities and temporally coherent surgical interactions.

    \subsubsection{Object-Level Persistence}
    We first establish object continuity across successive frames. Given two consecutive frames $t$ and $t+1$, candidate correspondences are restricted to object observations sharing the same component label (e.g., \textit{grasper}) and semantic type (e.g., \textit{instrument}). Their spatial compatibility is measured by the intersection-over-union (IoU) of their bounding boxes:
    \begin{equation}
    \operatorname{IoU}(b_a,b_b)
    =
    \frac{|b_a\cap b_b|}
    {|b_a\cup b_b|}.
    \label{eq}
    \end{equation}
    Among admissible candidates, a one-to-one assignment is obtained by maximizing the overall bounding-box overlap. Accepted correspondences introduce temporal coreference edges between matched observations, thereby linking otherwise isolated frame-level detections into persistent object tracks. These tracks provide the entity-level temporal continuity required for associating surgical interactions across frames.

    \subsubsection{Frame-level Action Grounding}
    Given the established object observations, we next resolve each semantic action triplet to the specific entities participating in the interaction. When a frame-level SG contains a verb-specific relation, its local object indices directly identify the corresponding instrument and target instances. This relation-guided grounding is particularly important when multiple instances share the same component label, such as two graspers appearing in the same frame, for which semantic labels alone are insufficient to determine which instance performs the action.

    If an endpoint specified by the semantic annotation is not supported by a visual detection, the missing endpoint is retained only as implicit graph evidence rather than instantiated as a reliable visual entity. This preserves the available semantic information while maintaining a distinction between visually observed and weakly inferred evidence. Each successfully grounded interaction is represented as a frame-level action edge connecting the participating entities.

    \subsubsection{Event-level temporal aggregation}
    Frame-level action edges are subsequently consolidated across time because a surgical maneuver typically persists over multiple consecutive frames. Rather than treating repeated occurrences of the same interaction as independent observations, compatible action edges are accumulated into an event state. For example, continuous gallbladder retraction across several frames is represented as a single temporally grounded event rather than as a sequence of redundant frame-level triplets.
    
    To accommodate transient occlusions and missed detections, an active event is allowed to remain open across short observation gaps. The event is terminated when no compatible interaction is observed beyond the predefined temporal tolerance or when the video ends. Each completed event records its participating entities, temporal boundaries, procedural phase context, and supporting frame-level evidence.
    
    Through these successive augmentations, the STSG preserves fine-grained frame-level observations while introducing two complementary forms of temporal structure: object tracks encode how surgical entities persist over time, whereas event nodes encode how their interactions unfold over temporal intervals. This representation therefore provides the structural evidence required to reason about when an interaction occurs, how long it persists, how different events are ordered or overlap, and how they relate to the surrounding object trajectories and procedural phases. Algorithm~\ref{alg:stsg_construction} summarizes the complete video-level STSG construction procedure.

    \begin{algorithm}[!t]
        \caption{Video-Level STSG Construction}
        \label{alg:stsg_construction}
        \begin{algorithmic}[1]
        \REQUIRE Frame-level SGs
        $\{\mathrm{SG}_{i,t}\}_{t\in\mathcal{T}_i}$,
        phase labels
        $\{p_i(t)\}_{t\in\mathcal{T}_i}$
        \ENSURE Video-level STSG $\mathrm{STSG}_i$
        
        \STATE Initialize $\mathrm{STSG}_i$, active object tracks, and open event states.
        \FOR{$t\in\mathcal{T}_i$ in temporal order}
            \STATE Add the frame objects, intra-frame relations, and phase context.
            \STATE Associate object observations with active tracks and add temporal coreference edges.
            \STATE Ground action triplets and insert frame-level action edges.
            \STATE Extend compatible event states or initialize new events.
            \STATE Close event states whose temporal tolerance has been exceeded.
        \ENDFOR
        \STATE Close the remaining events and attach their participants, temporal boundaries, and phase context.
        \RETURN $\mathrm{STSG}_i$
        \end{algorithmic}
    \end{algorithm}

\subsection{Evidence-Grounded Temporal Supervision}
\label{sec:temporal_qa_generation}

    Given the constructed $\mathrm{STSG}_i$, temporal QA pairs are generated by executing deterministic, phase-aware queries over graph events and object tracks. Each query specifies a semantic interaction, an answer scope, and the temporal property to be inferred. The interaction can be represented either as a complete triplet or as a partially specified interaction in which the instrument or target is left unspecified. This supports questions about both specific surgical interactions and broader patterns involving an action, instrument, or target. For comparison-based questions, we retain only fully grounded interactions with explicit instrument and target instances, excluding implicit endpoints that cannot be reliably verified from the video.

    Surgical phases serve as natural units for temporal sampling. For every individual phase, we implement early, middle, and late sampling windows across various temporal scales of up to 600 seconds. \textit{Boundary} and \textit{Phase-transition} questions instead use dedicated windows around adjacent phases. Importantly, the interval used to determine an answer is distinguished from the video segment presented to the model. The latter includes additional temporal context around the answer scope, whereas only evidence within the time interval explicitly stated in the question contributes to the answer. This design requires the model to localize the relevant interval while preventing surrounding context from contaminating the reference label.

    For each semantic query $q$, we collect all matching STSG events that overlap the answer interval $W$. The event intervals are clipped to $W$, and overlapping intervals or intervals separated by no more
    than $\delta$ are merged into a single temporal episode. We set $\delta=2$ seconds to reduce fragmentation caused by brief missed detections. The resulting episode set is denoted by
        \begin{equation}
            \mathcal{E}_{i,q}(W)
            =
            \left\{
            [u_m,v_m]
            \right\}_{m=1}^{M_{i,q}},
            \label{eq:temporal_episodes}
        \end{equation}
    where $M_{i,q}$ is the number of merged episodes. The corresponding episode count and cumulative duration are computed as
        \begin{equation}
            \begin{aligned}
                N_{i,q}(W)
                &=
                M_{i,q},\\
                D_{i,q}(W)
                &=
                \frac{1}{f_i}
                \sum_{m=1}^{M_{i,q}}
                \left(v_m-u_m+1\right),
            \end{aligned}
            \label{eq:temporal_count_duration}
        \end{equation}
    where $f_i$ denotes the temporal sampling rate, set to one frame per second (FPS) in our STSG construction. Consequently, \textit{Count} questions refer to distinct temporal episodes rather than individual event nodes, whereas \textit{Duration} questions measure the total length of the merged episode intervals.

    The same episode representation supports compositional temporal reasoning. \textit{Ordering} is determined from the first occurrence of each interaction, with temporally ambiguous comparisons excluded. \textit{Concurrency} is established from interval intersections between activities performed by different instruments, with a minimum sustained overlap used to suppress incidental frame-level coincidences. \textit{Extreme} questions identify the dominant instrument, target, or complete interaction according to cumulative duration or episode frequency; tied or insufficiently separated comparisons are discarded. Together, these categories require the model to reason about when an event occurs, whether two events coexist, and which activity predominates over an extended interval.

    Procedural annotations introduce two complementary forms of phase-aware reasoning. \textit{Boundary} questions focus on directly adjacent phase transitions using a symmetric 15 seconds window on each side of the boundary, clipped to the corresponding phase limits. They test whether the same interaction continues across the exact boundary, identify the first interaction, action, or target appearing afterward, and summarize the change from the last pre-boundary event to the first new post-boundary event. \textit{Phase-transition} questions operate at a broader scale by comparing up to the final 180 seconds of the preceding phase with the first 180 seconds of the subsequent phase. They characterize dominant early-phase actions and interactions, changes in dominant activity, the largest increase in episode frequency, and interactions newly visible after the transition. 

    Having defined the temporal relations underlying each question category, we construct negative instances from observed STSG facts rather than arbitrary semantic combinations. Zero-count queries use interactions absent from the selected scope but observed elsewhere in the video; \textit{Ordering} and \textit{Extreme} negatives reverse verified temporal or ranked relations; \textit{Concurrency} negatives pair co-visible but non-overlapping activities; and phase-aware negatives violate the corresponding boundary or transition conditions. This preserves semantic plausibility while requiring answers to be grounded in temporal evidence.

    After positive and negative candidates are instantiated, we apply a diversity-aware sampler that limits repeated use of the same semantic query and phase section; ensures coverage of all feasible question subtypes; and prioritizes candidates with unambiguous category-specific temporal evidence. As summarized in Table~\ref{tab:temporal_qa_distribution}, this procedure yields 18,458 QA pairs across 45 videos, with \textit{Concurrency} and \textit{Ordering} constituting the largest components.

        \begin{table}[!t]
        \caption{Distribution of temporal QA categories in STSG-VQA.}
        \label{tab:temporal_qa_distribution}
        \centering
        \footnotesize
        \renewcommand{\arraystretch}{1.08}
        \setlength{\tabcolsep}{2.5pt}
        \begin{tabularx}{\columnwidth}{lXrr}
        \hline
        \textbf{Category} &
        \textbf{Temporal focus} &
        \textbf{Number} &
        \textbf{Prop.} \\
        \hline
        Concurrency &
        Concurrent surgical activities &
        5,862 & 31.76\% \\
        Ordering &
        Temporal order of interactions &
        4,904 & 26.57\% \\
        Extreme &
        Longest or most frequent activity &
        2,752 & 14.91\% \\
        Count &
        Number of activity episodes &
        1,800 & 9.75\% \\
        
        Duration &
        Cumulative visible activity time &
        1,350 & 7.31\% \\
        Phase transition &
        Pre/post procedural change &
        1,197 & 6.48\% \\
        Boundary &
        Immediate cross-boundary behavior &
        593 & 3.21\% \\
        
        \hline
        \textbf{Total} &
        &
        \textbf{18,458} &
        \textbf{100.0\%} \\
        \hline
        \end{tabularx}
        \end{table}

    Finally, to mitigate template-induced shortcuts, we randomly sample from 108 distinct natural-language templates covering seven temporal-reasoning categories and their associated question subtypes. The resulting benchmark includes numeric, directional, binary, four-option multiple-choice, and concise open-ended answers. Each QA pair retains the displayed video segment, its exact answer scope, the supporting temporal segments, and the structured graph evidence used to derive the reference answer. Fig.~\ref{fig:STSG_with_2_QAs} (b) illustrates representative generated questions from the \textit{Concurrency} and \textit{Ordering} categories.

\section{Experiments}
\label{sec:experiments}

    Our experiments are organized around a central question: does structured temporal supervision improve a VLM’s ability to reason over the evolution of surgical events? We first compare zero-shot performance, static scene-graph supervision, and the proposed structured temporal supervision across seven temporal reasoning categories. We then test whether the resulting gains reflect increased use of visual evidence and whether structured temporal supervision transfers to and complements frame-level surgical VQA. Finally, we use STSG-VQA to characterize the remaining zero-shot limitations of current VLMs and the effect of increasing the frame budget.

\subsection{Dataset and Experimental Setting}
\label{subsec:dataset_protocol}

    The experiments use two complementary surgical VQA datasets: the proposed STSG-VQA benchmark and SSGVQA \cite{ssgvqa}. STSG-VQA is the primary benchmark for evaluating reasoning over temporally extended surgical events, whereas SSGVQA evaluates compositional understanding of individual surgical frames. This pairing allows us to separate the ability to recognize a surgical state from the ability to reason about how that state persists, changes, or interacts with other states over time.

    STSG-VQA is constructed from 45 laparoscopic cholecystectomy videos with aligned frame-level surgical scene graphs \cite{ssgvqa}, surgical phase annotations \cite{rendezvous}, and a manually verified set of valid frames. The videos span 90,089 seconds, corresponding to approximately 25.0 hours of operative video. After validity filtering and sampling at 1~FPS, the resulting STSGs contain 87,399 frame nodes, 634,460 object observations, and 27,143 temporally aggregated event nodes. To prevent visual and procedural leakage, the data are partitioned at the video level rather than at the QA level (Table~\ref{tab:dataset_split}).

    For the frame-level experiments, we follow the original SSGVQA formulation and evaluation protocol. SSGVQA contains question--answer pairs grounded in individual surgical frames and is therefore used both as a source of static scene-graph supervision and as a target for testing whether the benefits of structured temporal supervision extend to frame-level surgical VQA.

    \begin{table}[!t]
        \caption{Video-level partition and STSG-VQA statistics.}
        \label{tab:dataset_split}
        \centering
        \footnotesize
        \setlength{\tabcolsep}{3.2pt}
        \renewcommand{\arraystretch}{1.12}
        \resizebox{\columnwidth}{!}{
        \begin{tabular}{lrrrr}
        \hline
        \textbf{Split} &
        \textbf{Videos} &
        \textbf{Frames} &
        \textbf{Event nodes} &
        \textbf{QA pairs} \\
        \hline
        Training   & 35 & 65,535 & 20,180 & 13,932 \\
        Validation & 5  & 11,300 & 3,684  & 1,775  \\
        Test       & 5  & 10,564 & 3,279  & 2,751  \\
        \hline
        \textbf{Total} &
        \textbf{45} &
        \textbf{87,399} &
        \textbf{27,143} &
        \textbf{18,458} \\
        \hline
        \end{tabular}
        }
    \end{table}

\subsection{Evaluation Protocol}
\label{subsec:baseline_protocol}

    We evaluate three general-purpose open-source VLMs---Qwen2.5-VL-7B \cite{qwen25vl}, Qwen3-VL-4B \cite{qwen3vl}, and LLaVA-NeXT-Video-7B \cite{llavanextvideo}---together with the surgical-domain Hulu-Med-4B model \cite{hulumed}. For each QA item, frames are uniformly sampled in chronological order from the interval specified by \texttt{segment\_start\_time} and \texttt{segment\_end\_time}. Unless otherwise stated, a model receives 16 frames and a prompt containing the segment time span, natural-language question, required answer format, and output prefix \texttt{FINAL\_ANSWER:}. The answer scope is not used to crop the visual input. At inference, the model receives neither the reference answer nor the STSG-derived evidence used to generate and verify it, including the supporting event intervals, object tracks, and graph relations. Therefore, the observed fine-tuning gains cannot be attributed to direct access to the structured evidence from which the labels were derived.

    Predictions are aligned with references by \texttt{qa\_id}. Missing predictions are scored as incorrect, while duplicate and unmatched predictions are retained as diagnostics. Integer counts require an exact match after parsing the first integer. For duration questions, we use a continuous and non-decreasing tolerance that combines an absolute allowance for short intervals with a relative allowance for longer intervals:
    \begin{equation}
        \tau(r)=
        \begin{cases}
            3, & r\leq10,\\
            \max(3,0.20r), & 10<r\leq60,\\
            \min\!\left(30,\max(12,0.15r)\right), & r>60,
        \end{cases}
        \label{eq:duration_tolerance}
    \end{equation}
    where $r$ is the reference duration in seconds. The values from adjacent regimes agree at $r=10$ and $r=60$, avoiding an artificial change in scoring strictness at either boundary. A prediction $\hat r$ is correct when $|\hat r-r|\leq\tau(r)$. Binary answers are normalized across true/false and yes/no variants. Multiple-choice responses are parsed as A--D labels, with a fallback that maps predicted option text to the alternatives stated in the question. An unparseable deterministic answer is counted as incorrect.

    Open-ended responses are scored by a deterministic local Qwen3.5-4B judge \cite{qwen3.5}. The judge receives only the question, category and sub-category, segment windows, reference answer, and candidate answer, and returns a binary semantic-equivalence decision.

    Let $c_j\in\{0,1\}$ denote correctness for QA item $j$. We report question-level micro accuracy
    \begin{equation}
        \operatorname{Accuracy}_{\mathrm{micro}}
        =\frac{1}{N}\sum_{j=1}^{N}c_j,
        \label{eq:micro_accuracy}
    \end{equation}
    and define video-balanced accuracy as
    \begin{equation}
        \operatorname{Accuracy}_{\mathrm{video}}
        =\frac{1}{|\mathcal V|}
        \sum_{i\in\mathcal V}
        \frac{1}{N_i}\sum_{j\in\mathcal Q_i}c_j,
        \label{eq:video_balanced_accuracy}
    \end{equation}
    where $\mathcal Q_i$ contains the $N_i$ questions from video $i$. This second measure prevents videos with more generated questions from disproportionately determining the aggregate result.

    \begin{table*}[!t]
    \caption{Question-level accuracy (\%) across temporal categories in the 16-frame setting. $N$ denotes the number of test questions. ``+SSGVQA'' and ``+Ours'' denote fine-tuning on SSGVQA and STSG-VQA, respectively, followed by evaluation on the STSG-VQA test set. Parentheses show percentage-point changes from the corresponding zero-shot backbone.}
    \label{tab:category_results}
    \centering
    \footnotesize
    \setlength{\tabcolsep}{4.5pt}
    \renewcommand{\arraystretch}{1.12}

    \begin{tabular}{lccccccc}
    \hline\hline

    \textbf{Model} &
    \textbf{Concurrency} &
    \textbf{Ordering} &
    \textbf{Count} &
    \textbf{Extreme} &
    \textbf{Duration} &
    \textbf{Boundary} &
    \textbf{Phase trans.} \\

    $\boldsymbol{N}$ &
    1005 &
    758 &
    200 &
    421 &
    150 &
    69 &
    148 \\

    \hline

    Qwen2.5-VL &
    57.61 &
    47.89 &
    14.50 &
    38.95 &
    12.67 &
    42.03 &
    45.95 \\

    LLaVA-NeXT-Video &
    34.73 &
    33.91 &
    24.00 &
    26.60 &
    6.67 &
    37.68 &
    44.59 \\

    Qwen3-VL &
    53.13 &
    37.99 &
    19.50 &
    45.37 &
    10.67 &
    43.48 &
    47.97 \\

    Hulu-Med &
    51.44 &
    49.08 &
    20.50 &
    48.22 &
    14.67 &
    39.13 &
    47.30 \\

    Hulu-Med+SSGVQA &
    63.08 (+11.64) &
    50.40 (+1.32) &
    30.00 (+9.50) &
    47.51 (-0.71) &
    8.00 (-6.67) &
    55.07 (+15.94) &
    48.65 (+1.35) \\

    Hulu-Med+STSG-VQA (Ours) &
    80.40 (+28.96) &
    61.74 (+12.66) &
    \textbf{40.00 (+19.50)} &
    63.66 (+15.44) &
    16.67 (+2.00) &
    71.01 (+31.88) &
    62.16 (+14.86) \\

    Qwen3-VL+SSGVQA &
    57.91 (+4.78) &
    53.83 (+15.84) &
    32.00 (+12.50) &
    52.02 (+6.65) &
    7.33 (-3.34) &
    59.42 (+15.94) &
    41.22 (-6.75) \\

    Qwen3-VL+STSG-VQA (Ours) &
    \textbf{82.29 (+29.16)} &
    \textbf{63.19 (+25.20)} &
    34.00 (+14.50) &
    \textbf{68.17 (+22.80)} &
    \textbf{18.67 (+8.00)} &
    \textbf{72.46 (+28.98)} &
    \textbf{68.24 (+20.27)} \\

    \hline\hline
    \end{tabular}
    \end{table*}

\subsection{Implementation Details}
\label{subsec:implementation_details}

    We adapt Qwen3-VL-4B and Hulu-Med-4B to STSG-VQA using multimodal QLoRA. Both models are loaded with 4-bit NormalFloat quantization, double quantization, and \texttt{bfloat16} computation. LoRA modules are applied to language-attention, visual-to-language connector, and late vision-attention components, using component-specific ranks and learning rates fixed throughout training. The models are trained for 1.5 epochs with a per-device batch size of 1, 16 gradient-accumulation steps, and loss restricted to assistant-answer tokens. Training and the main test evaluation use 16 frames sampled uniformly and chronologically from each visual-context interval. Text-only evaluation is additionally used to quantify reliance on linguistic and procedural priors.

    We conduct three complementary evaluations involving SSGVQA. First, models fine-tuned only on SSGVQA are evaluated on the STSG-VQA test set; these rows are denoted by ``+SSGVQA'' in Table~\ref{tab:category_results}. They measure how far static scene-graph supervision transfers to temporal reasoning. Second, models fine-tuned only on STSG-VQA are evaluated directly on the official SSGVQA test split; these results test whether structured temporal supervision transfers to frame-level surgical VQA without direct SSGVQA fine-tuning. Third, SSGVQA-only and hybrid models are compared on the SSGVQA test split, where hybrid training combines SSGVQA and STSG-VQA examples. All experiments are executed on a single NVIDIA L40S GPU with 48~GB memory.

\subsection{Experimental Results}
\label{subsec:experimental_results}

\subsubsection{Structured Temporal Supervision Improves Temporal Reasoning}

    \begin{table}[!t]
    \caption{Micro accuracy (\%) under visual-input controls on STSG-VQA.}
    \label{tab:visual_evidence_ablation}
    \centering
    \footnotesize
    \setlength{\tabcolsep}{3.0pt}
    \renewcommand{\arraystretch}{1.10}
    \resizebox{\columnwidth}{!}{%
        \begin{tabular}{lccc}
            \hline
            \textbf{Model} &
            \textbf{Text} &
            \textbf{16 frames} &
            $\boldsymbol{\Delta}_{16-\mathrm{text}}$ \\
            \hline
            Qwen3-VL zero-shot
            & 41.26 & 42.49 & +1.23 \\
            Qwen3-VL + STSG-VQA
            & 49.98 & 66.88 & +16.90 \\
            \hline
            Hulu-Med zero-shot
            & 39.88 & 45.51 & +5.63 \\
            Hulu-Med + STSG-VQA
            & 51.58 & 65.07 & +13.49 \\
            \hline
        \end{tabular}%
    }
    \end{table}
    
    Table~\ref{tab:category_results} presents the central result of this study. Structured temporal supervision increases question-level micro accuracy from 42.49\% to 66.88\% for Qwen3-VL and from 45.51\% to 65.07\% for Hulu-Med, corresponding to improvements of 24.39 and 19.56 percentage points over their zero-shot backbones. It also outperforms static scene-graph supervision by 16.50 and 14.25 points, respectively. Crucially, these improvements extend across all seven temporal reasoning categories for both model families, whereas static scene-graph supervision produces uneven transfer and reduces duration accuracy for both models. This distinction shows that stronger recognition of instruments, targets, and within-frame relations is not sufficient for temporal reasoning. Learning from temporally connected surgical events additionally improves the model’s ability to reason about event persistence, concurrence, ordering, and procedural transitions.

    The largest gains over the zero-shot backbones occur in concurrency and boundary reasoning: Qwen3-VL improves by 29.16 and 28.98 points, while Hulu-Med improves by 28.96 and 31.88 points, respectively. These tasks require the model to relate simultaneous interactions or states on opposite sides of a procedural boundary, making their gains particularly consistent with learning event-level temporal structure rather than only a richer surgical vocabulary.

    The remaining weakness is also structured. Duration is the lowest-scoring category after STSG-VQA fine-tuning, reaching only 18.67\% for Qwen3-VL and 16.67\% for Hulu-Med; count reaches 34.00\% and 40.00\%, respectively. STSG supervision therefore improves the identification and comparison of event states more readily than exact temporal accumulation. This distinction suggests that learning which events coexist or change is easier than maintaining a calibrated memory of how often, or for how long, they occur under sparse video sampling.

    \begin{figure*}[!t]
    \centering
    \includegraphics[width=\textwidth]{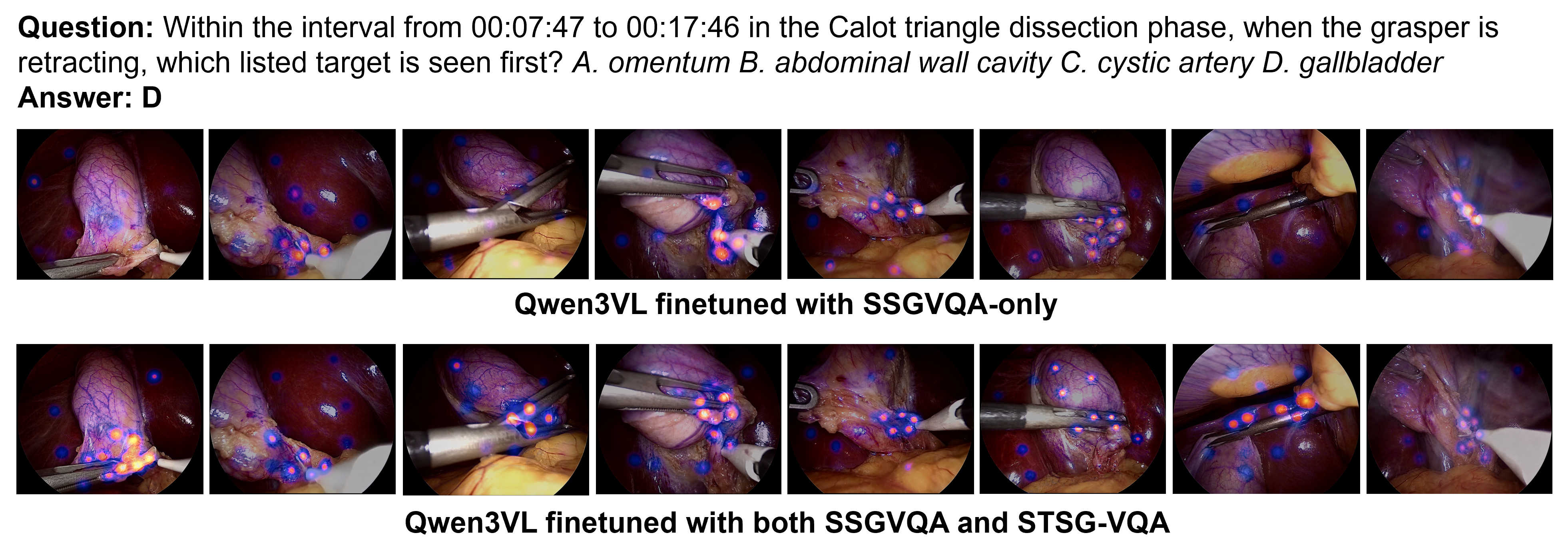}
    \caption{Decoder-attention comparison for the same temporal question. The top and bottom rows show Qwen3-VL-4B after SSGVQA-only and hybrid fine-tuning, respectively. Attention from answer-token positions to visual tokens is averaged over all heads, the final four decoder layers, and all answer-generation steps. Warmer colors indicate higher normalized attention.}
    \label{fig:attention_map}
    \end{figure*}

\subsubsection{Structured Temporal Supervision Increases the Contribution of Visual Evidence}

    Having established that structured temporal supervision improves temporal question answering, we next examine whether these gains reflect greater use of visual evidence rather than stronger reliance on linguistic or procedural priors. Table~\ref{tab:visual_evidence_ablation} tests whether the temporal gains can be explained by textual regularities alone. The zero-shot Qwen3-VL model already achieves 41.26\% without any image, only 1.23 points below its 16-frame result. Hulu-Med exhibits a larger but still limited 5.63-point visual advantage. These high text-only scores expose substantial answerability from question wording and procedural priors in both model families, and show why a high video-QA score alone is not sufficient evidence of visually grounded reasoning.

    STSG-VQA fine-tuning substantially changes the relationship between language and vision. The 16-frame advantage over text-only input expands to 16.90 points for Qwen3-VL and 13.49 points for Hulu-Med. Text-only accuracy also increases after fine-tuning, indicating that the supervision teaches useful surgical and task regularities; however, the much larger improvement when frames are present makes a purely linguistic-shortcut account insufficient. Because the graph evidence is not supplied at inference, the expanded visual margin is consistent with the model learning to map observed frame sequences to the event relations encoded by the STSG-derived targets. 

\subsubsection{Temporal Reasoning Complements Static Surgical VQA}

    \begin{table}[!t]
    \caption{Performance (\%) on the SSGVQA test set. ``STSG-VQA'' denotes direct evaluation after STSG-VQA-only fine-tuning; ``Hybrid'' combines SSGVQA and STSG-VQA supervision.}
    \label{tab:ssgvqa_raw_frame_baselines}
    \centering
    \footnotesize
    \setlength{\tabcolsep}{3.0pt}
    \renewcommand{\arraystretch}{1.12}
    \resizebox{\columnwidth}{!}{
    \begin{tabular}{lcccccc}
    \hline
    \textbf{Model} &
    \textbf{Setting} &
    \textbf{Acc.} &
    \textbf{mAP} &
    \textbf{mAR} &
    \textbf{mAF1} &
    \textbf{wF1} \\
    \hline
    Qwen3-VL-4B
        & zero-shot
        & 31.69
        & 17.14
        & 20.93
        & 16.70
        & 27.52 \\
    Qwen3-VL-4B
        & SSG-VQA
        & 67.18
        & 60.88
        & 54.56
        & 55.58
        & 66.61 \\
    Qwen3-VL-4B
        & STSG-VQA
        & 33.90
        & 22.49
        & 20.35
        & 17.77
        & 29.43 \\
    Qwen3-VL-4B
        & Hybrid
        & \textbf{71.01}
        & \textbf{65.07}
        & \textbf{57.86}
        & \textbf{59.09}
        & \textbf{70.60} \\
    \hdashline
    Hulu-Med-4B
        & zero-shot
        & 26.10
        & 20.70
        & 23.72
        & 17.32
        & 21.94 \\
    Hulu-Med-4B
        & SSG-VQA
        & 64.06
        & 54.72
        & 48.35
        & 49.00
        & 63.41 \\
    Hulu-Med-4B
        & STSG-VQA
        & 30.85
        & 21.93
        & 22.55
        & 17.57
        & 26.70 \\
    Hulu-Med-4B
        & Hybrid
        & 67.10
        & 56.11
        & 49.86
        & 50.47
        & 66.22 \\
    \hline
    \end{tabular}
    }

    \vspace{2pt}
    \begin{minipage}{\columnwidth}
        \scriptsize
        \textit{Note:} Acc. denotes overall accuracy. mAP, mAR, and mAF1 are the
        unweighted averages of per-class precision, recall, and F1 across the
        51 answer classes, respectively. wF1 denotes the class-support-weighted
        average of per-class F1.
    \end{minipage}
\end{table}

    We next test whether the benefits of structured temporal supervision extend beyond temporal question answering to frame-level surgical VQA. The two benchmarks examine complementary directions of transfer. On STSG-VQA, static scene-graph supervision improves aggregate accuracy for both backbones, but the resulting gains remain uneven across temporal reasoning categories (Table~\ref{tab:category_results}). Conversely, Table~\ref{tab:ssgvqa_raw_frame_baselines} shows that fine-tuning with structured temporal supervision alone improves SSGVQA accuracy from 31.69\% to 33.90\% for Qwen3-VL and from 26.10\% to 30.85\% for Hulu-Med. Weighted F1 also increases from 27.52\% to 29.43\% and from 21.94\% to 26.70\%, respectively. The transfer is not uniform across all metrics---mAR decreases slightly for both models---so temporal QA supervision does not replace direct frame-level training. Nevertheless, the improvements in accuracy, mAP, mAF1, and wF1 for both backbones indicate that learning event histories can reinforce reusable representations of static surgical interactions.

    Hybrid training provides the clearest evidence that the two supervisory signals are complementary. Relative to SSGVQA-only training, adding STSG-VQA examples improves every reported SSGVQA metric for both backbones. Qwen3-VL increases from 67.18\% to 71.01\% accuracy, with macro F1 rising from 55.58\% to 59.09\%; Hulu-Med increases from 64.06\% to 67.10\% accuracy, with macro F1 rising from 49.00\% to 50.47\%. Static scene graphs teach the model to resolve the constituents of a surgical state, whereas STSG-derived QA pairs additionally supervise persistence, co-occurrence, ordering, and transition. Their combination is therefore better understood as joint learning of surgical states and their dynamics, rather than as competition between two VQA datasets.

    \begin{table*}[!t]
    \caption{Zero-shot performance on the STSG-VQA test split. All models use the same prompting and evaluation protocol. Frames denotes the number of uniformly sampled visual observations.}
    \label{tab:baseline_results}
    \centering
    \footnotesize
    \setlength{\tabcolsep}{3pt}
    \renewcommand{\arraystretch}{1.18}
    \resizebox{\textwidth}{!}{
    \begin{tabular}{lcc|ccccc|cc}
    \hline\hline
    \textbf{Model} &
    \multicolumn{2}{c|}{\textbf{Setting}} &
    \multicolumn{5}{c|}{\textbf{Answer-Format Accuracy (\%)}} &
    \multicolumn{2}{c}{\textbf{Overall Accuracy (\%)}} \\
    \cline{2-10}
     &
    \textbf{Params} &
    \textbf{Frames} &
    \textbf{Binary} &
    \textbf{Count} &
    \textbf{Duration} &
    \textbf{Multi-choice} &
    \textbf{Open-ended} &
    \textbf{Video Avg.} &
    \textbf{Micro} \\
    \hline

    \multicolumn{10}{l}{\textit{Open-source general-purpose VLMs}} \\

    Qwen3-VL & 4B & 8 &
    40.85 & 19.00 & 10.67 & 49.87 & \textbf{28.30} & 39.37 & 40.79 \\

    Qwen3-VL & 4B & 16 &
    44.17 & 19.50 & 10.67 & 50.88 & 27.67 & 40.85 & 42.49 \\

    Qwen3-VL & 4B & 32 &
    49.38 & 18.50 & 10.67 & 51.73 & 25.79 & 43.23 & 44.67 \\

    Qwen2.5-VL & 7B & 8 &
    54.31 & 16.50 & 10.67 & 50.80 & 8.81 & 43.47 & 45.04 \\

    Qwen2.5-VL & 7B & 16 &
    55.55 & 14.50 & 12.67 & 51.14 & 6.29 & 43.66 & 45.47 \\

    Qwen2.5-VL & 7B & 32 &
    \textbf{56.59} & 15.50 & 8.67 & 50.72 & 7.55 & 43.79 & 45.62 \\

    LLaVA-NeXT-Video & 7B & 8 &
    48.06 & \textbf{28.50} & 7.33 & 24.85 & 5.03 & 31.59 & 31.92 \\

    LLaVA-NeXT-Video & 7B & 16 &
    48.06 & 24.00 & 6.67 & 24.77 & 5.66 & 31.23 & 31.55 \\

    LLaVA-NeXT-Video & 7B & 32 &
    48.06 & 24.00 & 9.33 & 24.77 & 6.29 & 31.32 & 31.73 \\

    \hline
    \multicolumn{10}{l}{\textit{Surgical VLMs}} \\

    Hulu-Med-4B & 4B & 8 &
    50.43 & 24.00 & \textbf{16.00} & 53.33 & 14.47 & \textbf{44.91} & \textbf{45.80} \\

    Hulu-Med-4B & 4B & 16 &
    50.14 & 20.50 & 14.67 & \textbf{53.58} & 15.09 & 44.39 & 45.51 \\

    Hulu-Med-4B & 4B & 32 &
    49.19 & 24.50 & \textbf{16.00} & \textbf{53.58} & 11.95 & 44.25 & 45.33 \\

    \hline\hline
    \end{tabular}
    }
    \vspace{2pt}

    \begin{minipage}{0.98\textwidth}
    \footnotesize
    \textit{Note:} Video Avg. denotes video-balanced accuracy.
    Micro denotes question-level accuracy over all test QA pairs.
    Open-ended answers are scored by the semantic-equivalence judge.
    \end{minipage}
    \end{table*}

\subsubsection{Qualitative Evidence of Temporally Selective Attention}

    We visualize decoder attention maps to show which visual regions receive attention while the fine-tuned Qwen3-VL-4B predicts its answer. Specifically, we extract causal self-attention from answer-token prediction positions to the visual tokens of eight uniformly sampled frames. We then average these weights across all attention heads, the final four decoder layers, and all answer-generation steps. The resulting per-frame visual-token scores are mapped back to their spatial patch grids, normalized, and interpolated to the original frame resolution for visualization. As shown in Fig.~\ref{fig:attention_map}, the SSGVQA-only model identifies salient anatomy around the Calot triangle and instrument--tissue interaction regions in individual frames, but does not consistently prioritize the evidence most relevant to the temporal-ordering question. Its attention therefore appears dominated by frame-level visual saliency. In contrast, the hybrid model exhibits attention that is qualitatively more concentrated on frames and regions associated with grasper retraction, suggesting improved alignment between the queried interaction and its occurrence across the sequence.

\subsubsection{Current VLMs Remain Limited in Zero-Shot Temporal Reasoning}

    After establishing the effect of the proposed structured temporal supervision, we use STSG-VQA to characterize the limitations of current general-purpose and surgical-domain VLMs under zero-shot evaluation. Table~\ref{tab:baseline_results} establishes the zero-shot performance landscape on STSG-VQA. The strongest overall result is obtained by Hulu-Med-4B with 8 frames, reaching 45.80\% micro accuracy and 44.91\% video-balanced accuracy. Under the common 16-frame setting, Hulu-Med and Qwen2.5-VL are effectively tied at 45.51\% and 45.47\% micro accuracy, respectively, followed by Qwen3-VL at 42.49\% and LLaVA-NeXT-Video at 31.55\%. Their video-balanced results follow the same general pattern. The negligible 0.04-point difference between Hulu-Med and Qwen2.5-VL indicates that surgical-domain specialization does not, by itself, confer a decisive advantage in fine-grained temporal reasoning.

    The frame-budget sweep further shows that access to more observations does not automatically produce stronger temporal reasoning. Qwen3-VL is the only model to improve consistently, increasing from 40.79\% with 8 frames to 44.67\% with 32 frames. In contrast, Qwen2.5-VL remains on a narrow plateau (45.04--45.62\%), LLaVA-NeXT-Video remains nearly unchanged (31.55--31.92\%), and Hulu-Med decreases slightly from 45.80\% to 45.33\%. Thus, temporal reasoning depends not only on the amount of visual evidence but also on whether the model can select, integrate, and retain information across observations.

    The answer-format results clarify what underlies these aggregate trends. Qwen2.5-VL is strongest on binary questions, Hulu-Med performs best on multiple-choice and duration questions, LLaVA-NeXT-Video obtains the highest count accuracy, and Qwen3-VL is strongest on open-ended answers. Notably, Qwen3-VL's gain from additional frames is driven primarily by binary accuracy, which rises from 40.85\% to 49.38\%, whereas its open-ended accuracy decreases from 28.30\% to 25.79\%. More frames can therefore improve coarse decisions without producing a corresponding improvement in semantically precise event reconstruction. Across all models, count, duration, and open-ended questions remain substantially more difficult than binary and multiple-choice decisions, reinforcing the need for supervision that explicitly models event evolution rather than relying on additional visual observations alone.

\section{Discussion}
\label{sec:discussion}

    The experiments distinguish surgical-state recognition from event-centered temporal reasoning. Zero-shot VLMs can often exploit linguistic and procedural priors, while static SSGVQA supervision improves instantaneous entity--relation recognition but transfers unevenly to temporal tasks. In contrast, STSG-VQA supervision improves all seven temporal categories, increases dependence on visual input, and complements static supervision in hybrid training. Because the STSG is used to construct supervision but is withheld at inference, these findings are consistent with VLMs internalizing structured knowledge about event persistence, concurrence, and transition rather than directly retrieving graph evidence.

    Several limitations constrain this interpretation. STSG-VQA contains 45 videos from a single procedure, and video-level splitting cannot establish generalization across institutions, acquisition systems, patient populations, or other procedures. Moreover, generated templates and phase-conditioned questions retain textual and workflow priors despite grounded negative construction and template diversification. External validation, held-out paraphrase families, phase-name masking, and temporally shuffled or reversed inputs would provide stricter tests of generalization and temporal dependence.

    Finally, 1-Hz STSG construction and uniform 16-frame sampling can miss brief interactions and repeated-event boundaries, contributing to the remaining difficulty in count and duration reasoning. Future work should investigate denser event-local sampling, uncertainty-aware tracking, hierarchical temporal memory, and explicit event-state updates. The present benchmark is intended for model development and evaluation and is not clinically validated for decision support.

\section{Conclusion}
\label{sec:conclusion}

    We introduced a structured temporal reasoning framework that augments frame-level surgical scene graphs with object-level continuity, event-level interaction continuity, and procedure-level connectivity. We then propose the STSG-VQA benchmark to supervise and evaluate seven forms of temporal reasoning. Zero-shot VLMs exhibit substantial reliance on textual and procedural priors, whereas structured temporal supervision improves every temporal category and substantially increases the contribution of visual evidence. Its transfer to frame-level SSGVQA, together with the consistent gains obtained through hybrid training, further shows that surgical-state recognition and temporal state-transition reasoning provide complementary supervisory signals. 
    
    These findings position structured temporal supervision not merely as a means to populate a benchmark, but as a reasoning scaffold that connects recognition of surgical states with modeling of their evolution. Duration estimation and event counting remain challenging, yet the resulting framework provides a concrete step toward surgical multimodal models that reason over persistent entities, interactions, and procedural transitions rather than isolated frames.

\end{document}